\documentclass[11pt,letterpaper]{article}
\usepackage{emnlp2017_camera_ready}
\usepackage{times}
\usepackage{latexsym}

\usepackage{graphicx}
\usepackage{comment}
\usepackage{url}
\usepackage{xspace}
\usepackage{CJKutf8}
\usepackage{tabu}

\emnlpfinalcopy

\def\emnlppaperid{***}

\newcommand{\system}[0]{MoodSwipe\xspace}

\newcommand{\ang}[0]{\textit{Anger}\xspace}
\newcommand{\joy}[0]{\textit{Joy}\xspace}
\newcommand{\sad}[0]{\textit{Sadness}\xspace}
\newcommand{\fear}[0]{\textit{Fear}\xspace}
\newcommand{\anti}[0]{\textit{Anticipation}\xspace}
\newcommand{\tired}[0]{\textit{Tired}\xspace}
\newcommand{\neu}[0]{\textit{Neutral}\xspace}

\newcommand{\SinicaAff}[0]{\ensuremath{1}\xspace}
\newcommand{\IMTAff}[0]{\ensuremath{2}\xspace}
\newcommand{\CMUAff}[0]{\ensuremath{3}\xspace}
\newcommand{\iet}[0]{\ensuremath{4}\xspace}

\title{\system: A Soft Keyboard that Suggests Messages\\Based on User-Specified Emotions}

\author{
Chieh-Yang Huang$^{\SinicaAff}$~~~
Tristan Labetoulle$^{\IMTAff}$~~~
Ting-Hao (Kenneth) Huang$^{\CMUAff}$\\
{\bf
Yi-Pei Chen$^{\SinicaAff}$~~~
Hung-Chen Chen$^{\SinicaAff}$~~~
Vallari Srivastava$^{\iet}$~~~
Lun-Wei Ku$^{\SinicaAff}$
}\\
$^{\SinicaAff}$ Academia Sinica, Taiwan.\\{\tt \{appleternity,ypc82,hankchen,lwku\}@iis.sinica.edu.tw}\\
$^{\IMTAff}$ IMT Atlantique, France. {\tt contact@tristan-labetoulle.com}\\
$^{\CMUAff}$ Carnegie Mellon University, USA. {\tt tinghaoh@cs.cmu.edu}\\
$^{\iet}$ Institute of Engineering \& Technology, DAVV, India. {\tt vallari357@gmail.com}
}

\date{}

\begin{document}

\maketitle

\begin{abstract}
We present \system, a soft keyboard that suggests text messages given the user-specified emotions utilizing the real dialog data. The aim of \system is to create a convenient user interface to enjoy the technology of emotion classification and text suggestion, and at the same time to collect labeled data automatically for developing more advanced technologies.
While users select the \system keyboard, they can type as usual but sense the emotion conveyed by their text and receive suggestions for their message as a benefit. In \system, the detected emotions serve as the medium for suggested texts, where viewing the latter is the incentive to correcting the former. We conduct several experiments to show the superiority of the emotion classification models trained on the dialog data, and further to verify good emotion cues are important context for text suggestion. 

\end{abstract}

\section{Introduction}
\label{sec:intro}
Knowing how and when to express emotion is a key component of
emotional intelligence~\cite{salovey1990emotional}.
Effective leaders are good at expressing
emotions~\cite{bachman1988nice};
expressing positive emotions in group activities improves
group cooperation, fairness, and overall group
performance~\cite{barsade1998group};
and expressing negative emotions can promote relationships~\cite{emotionRel2008}.
However, in the mobile device era where text-based communication is
part of life,
technologies are rarely utilized to assist users to \textit{express}
their emotions properly via text.
For instance, business people in heated disputes with their clients may
need assistance to rephrase their angry messages into neutral descriptions
before sending them.
Similarly, people may have trouble finding the perfect words to show how much 
they appreciate a friend's help.
Or people may want to deliberately express anger to
extract concessions in negotiations, or to make a joke, such as with the
``Obama's Anger Translator'' skit, in which the comedian ``translates'' the
U.S. President's calm statements into emotional
tirades.
While emotion classification has been used in helping users to better
understand other people's emotions~\cite{wang2016sensing,huang2017challenges},
these technologies have rarely been used to support user needs in
expressing emotions.
Most prior work focuses on interface design, for instance using kinetic typography
or dynamic
text~\cite{bodine2003kinetic,forlizzi2003kinedit, Lee:2006:UKT:1142405.1142414}
,
affective buttons~\cite{Broekens:ACII09},
or text color and emoticons~\cite{Sanchez:IHC06}
to enable emotion expression in instant messengers.
Other work explores the relations between user typing patterns and
their emotions~\citep{zimmermann2003affective,alepis2006affective,keymochi}.
However, the text itself is still the essential part of text-based
communication; visual cues and typing patterns are only minor factors.
Other work even describes attempts to incorporate body signals such as
fluctuating skin conductivity levels~\cite{dimicco2002conductive}, thermal
feedback~\cite{wilson2016hot}, or facial expressions~\cite{el2004faim} to enrich
emotion expression, but these require additional equipment and
are less scalable.

\begin{figure*}[t]
    \small
    \centering
    \includegraphics[width=0.95\textwidth]{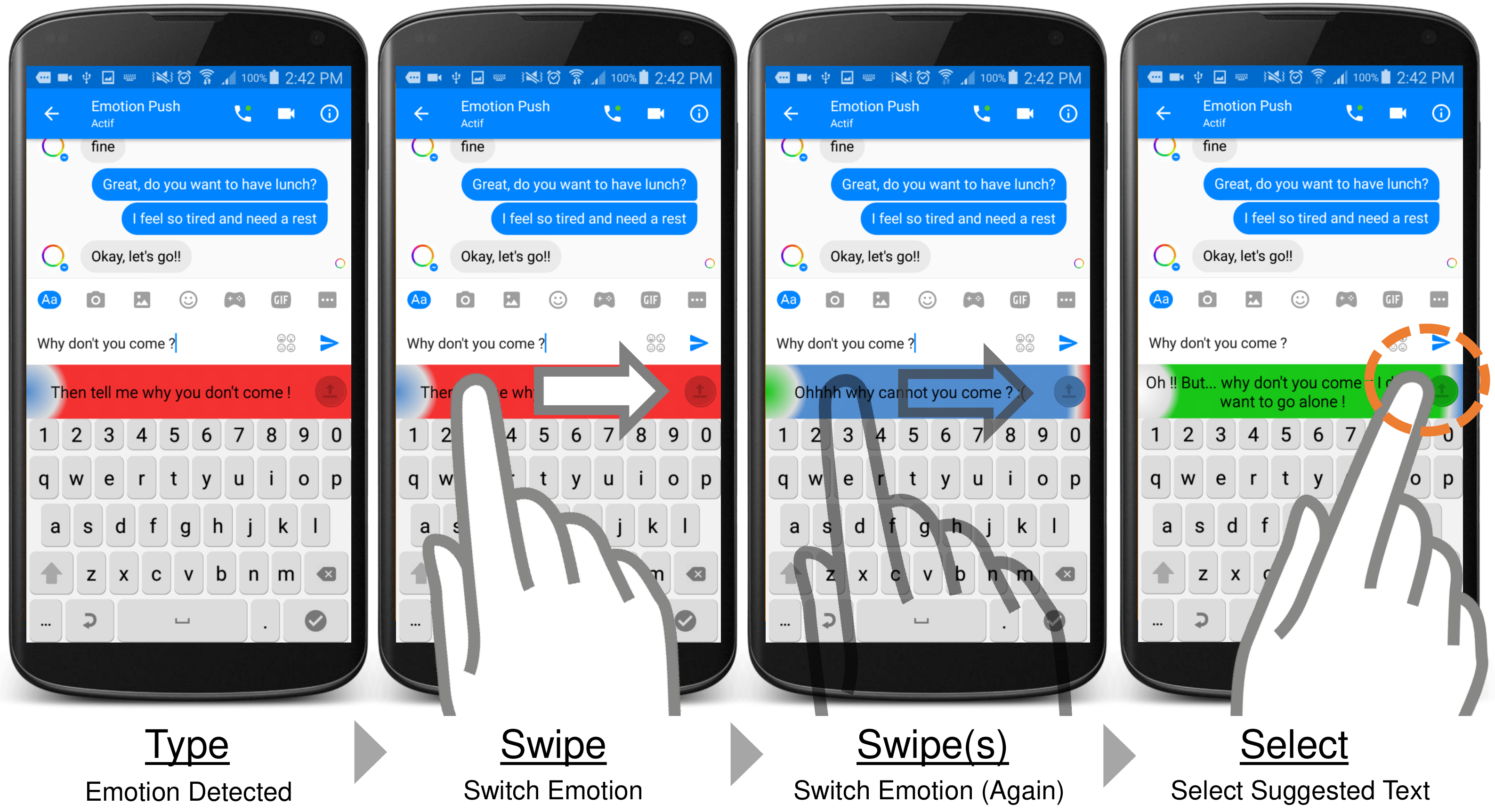}
	\vspace{-.8pc}
	 \caption{The user interface of \system keyboard, which includes a standard soft keyboard, a color bar above the keyboard, and a circle button to the
right of the color bar. Users swipe the color bar to specify their emotions and view the messages suggested for different emotions.}
    \label{fig:interface}
	\vspace{-.5pc}
    
\end{figure*}

In this paper we introduce \textbf{{\em \system}}\footnote{\system is available at:  \url{https://play.google.com/store/apps/details?id=sinica.moodswipe&hl=en}}, a soft keyboard that automatically
suggests text according to user-specified emotions.
As shown in Figure~\ref{fig:interface}, \system receives user input text from
the keyboard, and immediately shows the detected emotion of the input message.
Seven emotions (\ang, \joy, \sad, \fear, \anti, \tired and \neu) are
detected and presented by their corresponding colors as defined by the research of
psychologists and user studies~\cite{wang2016sensing}.
Users \textbf{\textit{swipe}} the color bar to change the detected emotion
according to their \textbf{\textit{mood}} and view the messages suggested for
different emotions.
For example, if the user types ``I disagree,'' \system suggests a
relevant message telling the user how one would say the same thing when he or she
is happy, sad, or angry.

The contributions of this work are three-fold.
First, we address the long-standing challenge of \textbf{collecting
\textit{self-reported} emotion labels for dialog messages}.
Unlike posts on social media, where users often spontaneously tag their own
emotions, self-reported emotion labels for dialog messages are
expensive to collect.
Users often feel disturbed when they are asked to annotate their own emotions
on the fly.
\system provides a handy service which is entertaining and easy to use.
It provides a natural incentive for users to label their own emotions on the spot.
Second, \system closes the loop of \textbf{bi-directional \textit{interactive
emotion sensing}}.
Most prior work powered by emotion detection focused on helping
users when \textit{receiving} messages~\cite{wang2016sensing} instead of when
\textit{sending} them.
\system enables automated support, helping users express their emotions in text,
and therefore supplies the missing piece for an
emotion-sensitive text-based communication environment.
Finally, \system introduces a \textbf{new interaction paradigm}, in which users
explicitly provide feedback to systems about \textit{why} they select this
suggested response.
Classic response suggestion tasks such as dialog
generation~\cite{li2016deep} or automated email
reply~\cite{kannan2016smart} assume that the in-the-moment context of each user
(e.g., the current emotion) is unknown.
\system opens up possibilities for users to explicitly and actively provide
context on the fly, which the automated models can use to provide better
suggestions.

\begin{figure}[t]
    \small
    \centering
    \includegraphics[trim=4cm 10cm 1.5cm 3cm, width=\columnwidth]{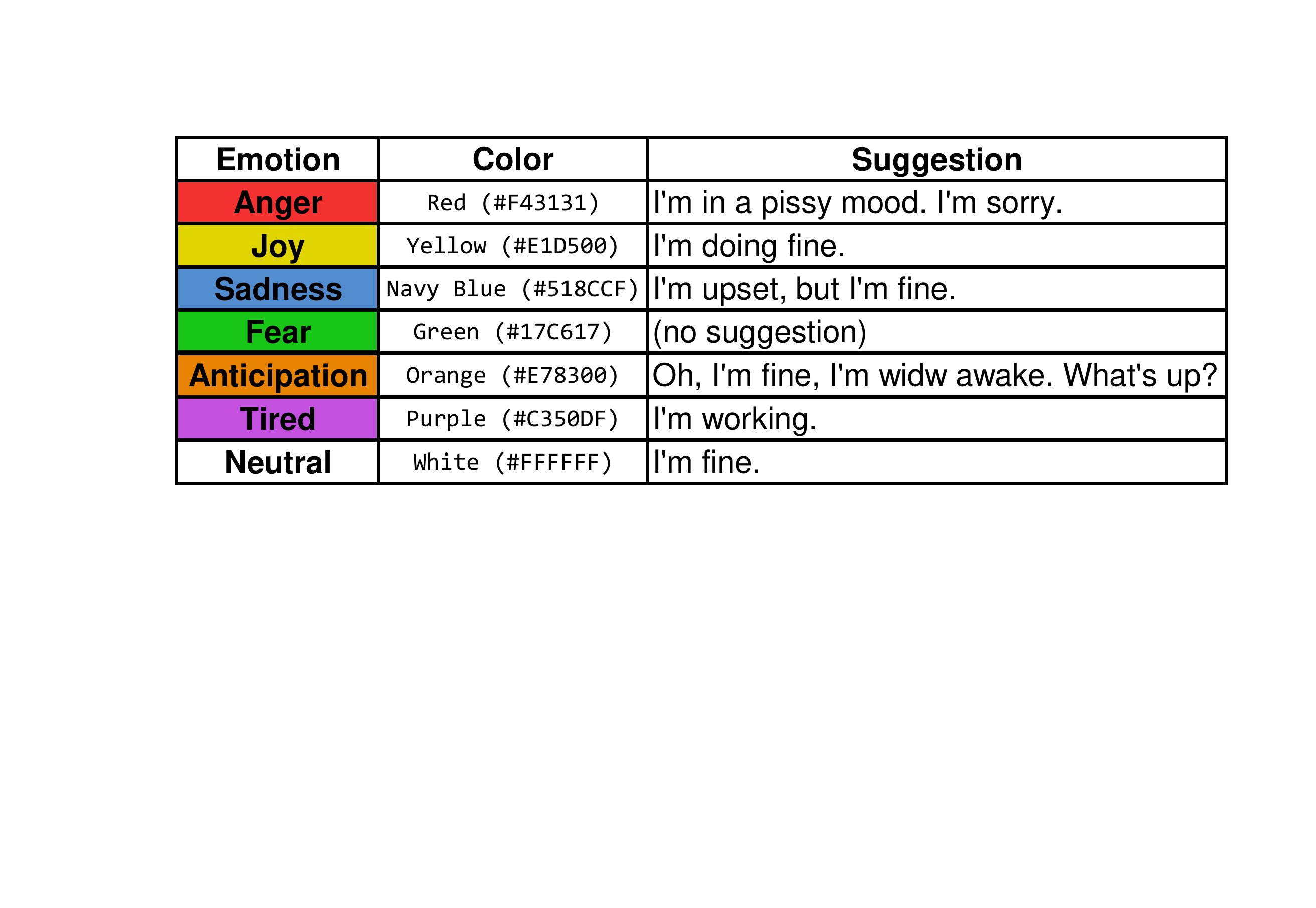}
    \vspace{-.5pc}
    \caption{Mapping between colors and emotions, and example suggestion texts for ``I am fine.''}
    \label{fig:colormap}
    \vspace{-.8pc}
\end{figure}

\section{The \system Keyboard}



The user interface and workflow of \system are shown in Figure ~\ref{fig:interface}. 
The \system keyboard interface contains three major parts: \emph{(i)} a standard soft keyboard, \emph{(ii)}
a color bar above the keyboard, and \emph{(iii)} a circle button to the
right of the color bar.
When the user starts typing, \system detects the emotion of the input text in
real time, and the color bar background changes color on-the-fly to reflect
the current emotion of the user input text.
\system's seven emotions and their colors are
shown in Figure~\ref{fig:colormap}, which was developed based on 
psychological work and user studies~\cite{wang2016sensing}.
Based on the emotion \system currently
displays, the color bar also shows the user the suggested text. Those suggested messages for the input ``I am fine'' are also listed in Figure~\ref{fig:colormap}.

When the user \textbf{\textit{swipes}} the color bar, one of \system's seven emotion colors 
is brought up in descending order of the classification probability
for the input message. 
The user swipes right to see 
the predicted emotion with a lower probability
and its suggested text, or swipes left to see the previous one.
In Figure~\ref{fig:interface}, the user types ``Why don't you come?'' and \system detects \ang (red) 
and suggests ``Then tell me why you don't come!''.
The user swipes right to see the option ``Ohhhh why cannot you come?'' for emotion \sad (blue). The user keeps swiping until he or she reaches a suitable message (``Oh!! But...why don't you come I don't want to go alone!'') and then clicks the circle button at color bar's right side to replaces the user input with the suggested text. Then with this replacement, the user self-reports that the emotion label of the user's message ``Why don't you come?'' should be more \fear (green) than \ang (red) in the current dialog context.


 
\system actively triggers emotion detection and updates color accordingly when the user presses the spacebar, which usually indicates the completion of a
word, or has a 500ms pause after the last user input.
To lighten server loads and reduce possible conflicts with the second condition, the minimum time interval between two triggers is set at 400ms.
All users activities are recorded, including typed text, suggested texts, emotion labels selected/abandoned, and all the timestamps.

\section{Use Cases}

In this section we outline several possible use cases of \system.
First, \system can help users to better \textbf{understand their own messages' emotions} perceived by other users.
Our prior study~\cite{huang2017challenges} shows that not all users are clearly aware of what emotion their messages will convey.
In this scenario, \system is able to act as an early assistant or reminder when composing a message.
Second, \system can \textbf{assist users to better express themselves}
when ``words fail me.''
Sometimes users could experience strong emotions and have difficulty in finding good ways to express themselves via text, and they can
type keywords into \system to search for better messages from its dialog database.
Third, users can \textbf{alternate the perceived emotions in their own texts} for various purposes.
For instance,
some people might need assistance to rephrase their angry messages into neutral descriptions, and some people may want to deliberately express anger to
extract concessions in negotiations.
Finally, \system can be used as a tool to help new users quickly \textbf{adapt to the language style of a community.}
\system is powered by messages that were collected from young IM users.
An elder new user who is not very familiar with the language styles of the young generation can use \system to rephrase his/her sentence so that it can be better received by young users.
\section{Back-end System, Experiments and Discussions}

Two major functions of the \system keyboard are to guess for the users the emotion
of their current text message and to provide text suggestions based on that. In
this section, we describe several models we developed and different settings
used to evaluate their performance. Our advantage in conducting these experiments
comes from our emotion-based chat app 
EmotionPush~\cite{wang2016sensing} and the social dialogs it has collected. 

\subsection{Experimental Materials}
\label{sec:data}

For the experiments, we adopted the EmotionPush dataset (available soon). A total number of 162,031
message logs were collected for this dataset. To evaluate the performance of
emotion classification, we had native speakers manually label the emotions of the randomly selected
8,818 messages. These manual emotion attribute were all chosen from the
seven emotion labels defined for the keyboard. Among these 8,818
emotion labeled messages, 70\% were for training, 10\% for validation and
20\% for testing. 
Two different emotion corpus, LJ40K \cite{yang2013quantitative} and the tweet data, are utilized for comparison.
LJ40K contains 40 emotions and for each of the emotions, 1,000 blogs are collected. The 40 emotions are then mapped to the 7 emotions according to our previous work~\cite{wang2016sensing}.
On the other hand, the tweet data is built by Twitter streaming API\footnote{\url{https://dev.twitter.com/streaming/overview}} with a filter of using the 40 emotions as hashtag. A total number of 19,480 tweets are collected and further categorized into 7 emotions.

The text message suggestion function provided by \system recommends responses
given the input text message and its corresponding emotion label from users. For the
experiments, we selected messages from the labeled 8,818 messages according to two
rules: 1) to ensure a proper turn, the previous message must be sent from another 
user instead of the same message owner, and 2) the emotion label must not be neutral.
We drop a small number of short messages containing hindi or pure punctuation 
(e.g., ``!!!'') for which text suggestions cannot be found, as in these cases we are
unable to evaluate the performance of different settings. A total of 1,366 messages were collected for the sentence suggestion experiment (707
Joy, 223 Anger, 189 Sadness, 124 Anticipation, 72 Fear and 51 Tired).
For the evaluation, text suggestions are generated for these 
messages using \system.

\subsection{Emotion Classification}

Two models are developed for emotion classification: the general
CNN~\cite{kim2014convolutional} with 125 filters, including 25 filters for each
filter length ranging from 1 to 5, and the 
LSTM~\cite{hochreiter1997long}. These two models are trained on blog data,
tweet data, and our dialog data, respectively, and then tested on the dialog
data. 
In Table~\ref{table:emotion_prediction_result} we report the results of three 
major emotions with these two models, as the other emotions are minor and
the training data insufficient to build a reliable model
(anticipation 1.77\%, tired 0.8\%, fear 1.19\%, total 3.77\%). Only accuracies trained on the dialog data for three major emotions are all over 0.9 (see CNN$^3$ and LSTM$^3$), which supports the use of dialogs in \system. Considering time-consuming issue, we adopted CNN as the final model for \system. 
\begin{table}[t]
    \centering
    \small
    \resizebox{\columnwidth}{!}{
        \begin{tabular} {|l|r|r|r|r|}
        \hline
        \textbf{Model} & \textbf{Joy} & \textbf{Anger} & \textbf{Sadness} & \textbf{Neutral} \\ \hline
        \cite{wang2016sensing}        & .779 & .771 & .853 & .323 \\    
        CNN$^1$         & .832 & .960 & .750 & .513 \\
        CNN$^2$         & .645 & .942 & .503 & .222 \\
        CNN$^3$         & .905 & .962 & .973 & .820 \\
        LSTM$^1$        & .230 & .967 & .963 & .222 \\
        LSTM$^2$        & .596 & .959 & .516 & .222 \\
        LSTM$^3$        & .906 & .965 & .964 & .816 \\
        
        \hline
        \end{tabular}
    }
    \vspace{-.2cm}
    \caption{Accuracy of the emotion classification task tested on dialog data while trained on blog$^1$, tweet$^2$ and dialog$^3$ data.}
    \label{table:emotion_prediction_result}
\end{table}

\subsection{Text Suggestion \& Results}


The purpose of the experiments for test message suggestion is to determine whether
the system generates better suggested texts given the user-specified emotion. We 
designed a retrieval-based model utilizing Lucene~\cite{McCandless:2010:LAS:1893016} 
and then applied it on the EmotionPush dataset which contains 162,031 social dialog
messages. 
When searching for similar messages, Lucene first applies term matching on the dataset using query message. Messages containing at least one same word would be candidates which is then ranked by BM25~\cite{robertson2009probabilistic}.
When the user received a message and is composing a response, the user manually specifies an emotion ({\em e.g.}, \textit{Anger}) that he/she wants to convey in the message, and the following two settings for generating responses.

\begin{enumerate}

\item \textbf{[Baseline]: } 
Given the message that the user received, \system first retrieves its most similar message (by Lucene) from the database, and then returns the response of that retrieved message as the suggestion.
\item \textbf{[+Emotion]:} The procedure is identical as [Baseline], just that the suggested text must convey the user-specified emotion.
For instance, if the user specifies \textit{Anger}, \system takes the message that the user received and from database finds its most similar message whose response's emotion is labelled as \textit{Anger} as the suggestion.

\end{enumerate}

Note that the emotions in our database are annotated by automatic algorithms instead of humans.

To assess the quality of suggested messages in each setting, we use the 1,366 messages collected in sentence suggestion experiment (Section~\ref{sec:data}) to conduct human evaluations with crowd workers recruited via Amazon Mechanical Turk.
For each message, we first show its 10 previous messages in the original chat log to crowd workers.
We then show the following three messages, in a random order, as the \textit{follow-up line} candidates of the displayed chat log: 1) the actual user input response, 2) the suggested texts in [Baseline], and 3) the suggested texts in [+Emotion]. 
Workers are asked to rank these three candidate messages based on their \textit{clarity}, \textit{comfort}, and \textit{responsiveness} of being the follow-up line of the given chat log~\cite{liu2010quality} ($rank=1,~2,or~3$. Lower is better.)
For each message, we recruit 5 workers and average their results.
Table~\ref{table:turker_rank} shows the average ranking and the ``Good Suggestion Rate'' of each setting, which is the proportion of the suggested messages that have a better (lower) ranking than the original user input response.
While the original input responses still have a better (lower) average ranking, Table~\ref{table:turker_rank} shows that 26\% to 28\% of the suggested texts are considered good by crowd workers and thus could be useful to users.
Results also show that the [+Emotion] setting on average generates slightly better suggestions than [Baseline] setting in all three aspects.

With the consideration that in the three evaluation aspects \textit{comfort} is most relevant to emotions,
we further analyze the \textit{comfort} result of each emotion (Table~\ref{table:comfort_result}.) 
Among all emotions, the Good Suggestion Rates for \emph{Anger} messages are the highest (40.36\% and 37.49\%), which are even much higher than the average rates (about 28\% as shown in Table~\ref{table:turker_rank}).
This result suggests that our method is particularly useful for expressing Anger emotions.


\begin{table}[t]
\centering
\small
\begin{tabular}{|c|c|c|c|}
\hline
Setting                         & Clarity & Comfort & Responsiveness \\ \hline
\multicolumn{4}{|c|}{Rank of Messages and Suggested Texts}                                           \\ \hline
Input                                   & 1.522  & 1.570  & 1.531         \\ \hline
\textbf{Baseline}                       & 2.245  & 2.220  & 2.244         \\ \hline
\textbf{+Emotion}                       & 2.233  & 2.210  & 2.225         \\ \hline
\multicolumn{4}{|c|}{Good Suggestion Rate (\%)}                         \\ \hline
\multicolumn{1}{|c|}{\textbf{Baseline}} & 26.12   & 28.38   & 26.44          \\ \hline
\multicolumn{1}{|c|}{\textbf{+Emotion}} & 26.09   & 28.65   & 26.70          \\ \hline
\end{tabular}
\caption{Human evaluation results. In Baseline the user-specified emotions are not available.}
\label{table:turker_rank}
\end{table}

\begin{table}[t]
\centering
\small
\begin{tabular}{|c|c|c|l|}
\hline
\multicolumn{4}{|c|}{Good Suggestion Rate (\%)} \\ \hline
Setting & Anger & Anticipation & Fear  \\ \hline
\textbf{Baseline}       & 40.36 & 21.29        & 31.39 \\ \hline
\textbf{+Emotion}       & 37.49 & 20.32        & 25.28 \\ \hline 
Setting & Joy   & Sadness      & Tired \\ \hline
\textbf{Baseline}       & 25.35 & 29.31        & 27.45 \\ \hline
\textbf{+Emotion}       & 28.18 & 26.56        & 29.41 \\ \hline
\end{tabular}
\caption{Good suggestion rates of \textit{comfort} for messages of different emotions.}
\label{table:comfort_result}

\end{table}

These results show the potential benefits of including emotion signals in a response suggestion application.
While the simple retrieval-based model (where emotions act only as ``filters'') may be of limited use, \system is still able to suggest responses that are better than the user responses around 25\% of time.
We believe that when \system is deployed, more data can be collected and a more sophisticated models can be developed to
boost the benefit of emotion context.










\subsection{Collecting User-reported Labels}
One merit of \system is the capability to collect user-reported labels.
\system can obtain labels from two major user actions, select and swipe, respectively.

\begin{enumerate}

    \item \textbf{[Select]} When the user first types a response, browses all suggestions, and finally selects one suggested text, \system can record the user's original typed text and label its emotion as that of the selected text.
     
    \item \textbf{[Swipe]} When the user first types a response, swipes directly to a specific emotion ({\em e.g.}, Joy) and stops there, even without selecting the suggested text, it often still indicates that the user wants to express this emotion ({\em e.g.}, Joy.) Therefore, \system can record the user's current typed text and label it as the same emotion, even the user does not select the suggested text eventually.

    

\end{enumerate}

\section{Conclusion}

We have developed the sender side \system to cooperate with the receiver side applications and complete the emotion sensitive communication framework. \system provided a convenient interface which facilitating users on using the modern emotion classification and text suggestion techniques. In \system, data are labeled automatically according to frond-end cues in the background. We show that the user-specified emotion can benefit text suggestion, though the performance can still be improved by increasing the size of the dialog database.
\system is available at Google Play, and a demo video is provided at: \url{https://www.youtube.com/watch?v=SZ1biWoiq3Y}

\section*{Acknowledgments}
Research of this paper was partially supported by Ministry of Science and Technology, Taiwan, under the contract 105-2221-E-001-007-MY3.

\bibliography{emnlp2017_camera_ready}
\bibliographystyle{emnlp_natbib_camera_ready}

\end{document}